\pdfoutput=1
\documentclass[letterpaper, 10 pt, conference]{ieeeconf}  % Comment this line out if you need a4paper

\IEEEoverridecommandlockouts                              % This command is only needed if 
\usepackage{amsmath} % assumes amsmath package installed
\usepackage{amssymb}  % assumes amsmath package installed
\usepackage{float}
\usepackage{tabularx}
\usepackage{graphicx}
\usepackage[capitalise]{cleveref}
\usepackage{color, colortbl}
\usepackage{hhline}
\usepackage{multicol}
\usepackage{subfigure}
\usepackage{fancyhdr}
\usepackage{multirow}
 
\title{\LARGE \bf
Attention-based Lane Change Prediction
}

\author{Oliver Scheel$^{1, 2,*}$, Naveen Shankar Nagaraja$^{1,*}$, Loren Schwarz$^1$, Nassir Navab$^2$, Federico Tombari$^2$% <-this % stops a space
\thanks{* indicates equal contribution}
\thanks{$^1$Oliver Scheel, Naveen Shankar Nagaraja and Loren Schwarz are with BMW Group, 80788 M\"unchen, Germany, {\tt\small \{oliver.scheel, naveen-shankar.nagaraja, loren.schwarz\}@bmw.de}}%
\thanks{$^2$Federico Tombari and Nassir Navab are with the Faculty of Computer Science,
	Technische Universit\"at M\"unchen, 85748 Garching bei M\"unchen, Germany, {\tt\small \{tombari, navab\}@in.tum.de}}%
}

\begin{document}

\setlength{\textfloatsep}{5pt plus 1.0pt minus 2.3pt}

\maketitle
\thispagestyle{fancy}
\pagestyle{empty}

%%%%%%%%%%%%%%%%%%%%%%%%%%%%%%%%%%%%%%%%%%%%%%%%%%%%%%%%%%%%%%%%%%%%%%%%%%%%%%%%
\begin{abstract}
Lane change prediction of surrounding vehicles is a key building block of path planning. The focus has been on increasing the accuracy of prediction by posing it purely as a function estimation problem at the cost of model understandability. However, the efficacy of any lane change prediction model can be improved when both corner and failure cases are humanly understandable. We propose an attention-based recurrent model to tackle both understandability and prediction quality. We also propose metrics which reflect the discomfort felt by the driver. We show encouraging results on a publicly available dataset and proprietary fleet data.
\end{abstract}

%%%%%%%%%%%%%%%%%%%%%%%%%%%%%%%%%%%%%%%%%%%%%%%%%%%%%%%%%%%%%%%%%%%%%%%%%%%%%%%%

\section{INTRODUCTION}
Artificial intelligence is commonly seen as the key enabler for fully autonomous driving. Sensing and Mapping, Perception, and (Path) Planning are often seen as the building blocks of any non-end-to-end autonomous system. The rise of deep learning has led to an unprecedented progress in Mapping and Perception. However, path planning has a hybrid nature - it tends to be model-driven with some sub-components learned using deep learning. This is primarily due to the severely complex interaction of different agents (static and dynamic) and prior knowledge (map and traffic information). Dearth of data which includes various corner cases further limits completely data-driven based planning.

Prediction is a crucial part of autonomous driving, serving as a `lego block' for tasks like Path Planning, Adaptive Cruise Control, Side Collision Warning, etc.
%Prediction itself can be further categorized into prediction of trajectories or maneuvers of other vehicles on the road, the behavior prediction of vulnerable road users (pedestrians, cyclists), to name a few.
In this work, we address the problem of predicting lane changes of vehicles. This is of paramount importance, as around 18\% of all accidents happen during lane change maneuvers \cite{tsa}, and lane changes are often executed in high-velocity situations, e.g. on highways. A precise prediction thus decreases risk and enables safer driving. This safety gain stemming from a sensitive prediction is one side of the coin. On the other hand, though, false  predictions have to be avoided as they have a negative influence on driver comfort. Each false prediction results in unnecessary braking or acceleration.

\begin{figure}[!t]
\centering
\includegraphics[scale=0.6]{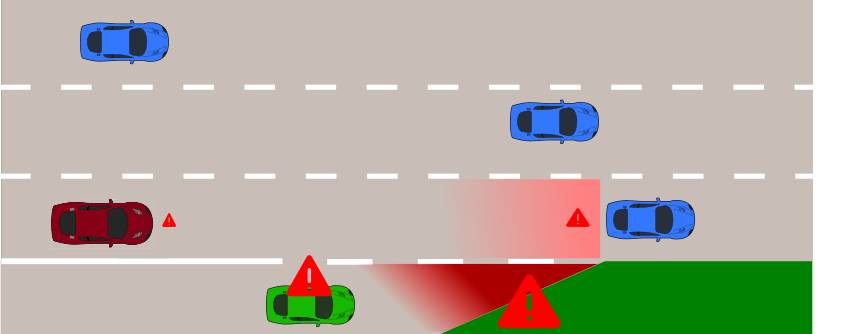}
\caption{Sample image of how an attention mechanism perceives a scene: when predicting the imminent lane change of the target (green car), a strong weight is given to the ending on-ramp. Furthermore, intrinsic features of the target, like lateral velocity, also get a high weight - as they are good indicators in general. A small weight is given to the neighboring cars in the adjacent lane - the gap is determined not critical for this merging maneuver. The ego car (in red) thus slows down smoothly.}
\label{fig:teaser}
\end{figure}

For predicting lane changes, several ``classical" models, like Support Vector Machines (SVMs) \cite{potsvm} or Random Forests \cite{DaimlerRF}, have been proposed, with only one recurrent neural net having been published recently \cite{toyotasrnn}. These classical methods, though theoretically sound, see maneuver prediction as function estimation. Though the weights on different features can give us a hint as to what the function considers important, understanding these models when prior knowledge is also given as input has lacked clarity in analysis. The question we ponder over is: \emph{does/can a system see what a human looks at?}, e.g. when one approaches a highway entry ramp the probability of a lane change for vehicles on the ramp is higher, and the human driver slows down with this prior knowledge (see \cref{fig:teaser}).

To answer the above intriguing question, we:\\
\noindent\textbf{(a)}~propose the first recurrent neural network making use of an attention mechanism over different features and time steps. This model is designed to understand complex situations and also explain its decisions. Like humans it can shift its focus towards certain important aspects of the current scene.

\noindent\textbf{(b)}~introduce metrics which indirectly reflect driver's comfort, and thus allow a meaningful quantification of prediction quality. % \todo{explain better what you mean by reflecting driver's comfort}

\noindent\textbf{(c)}~provide the first comprehensive evaluation of several models aimed at the same task on the same benchmark, and analyze critical corner cases and visually interpret them.

\begin{figure}[!h]
\centering
\includegraphics[scale=0.5]{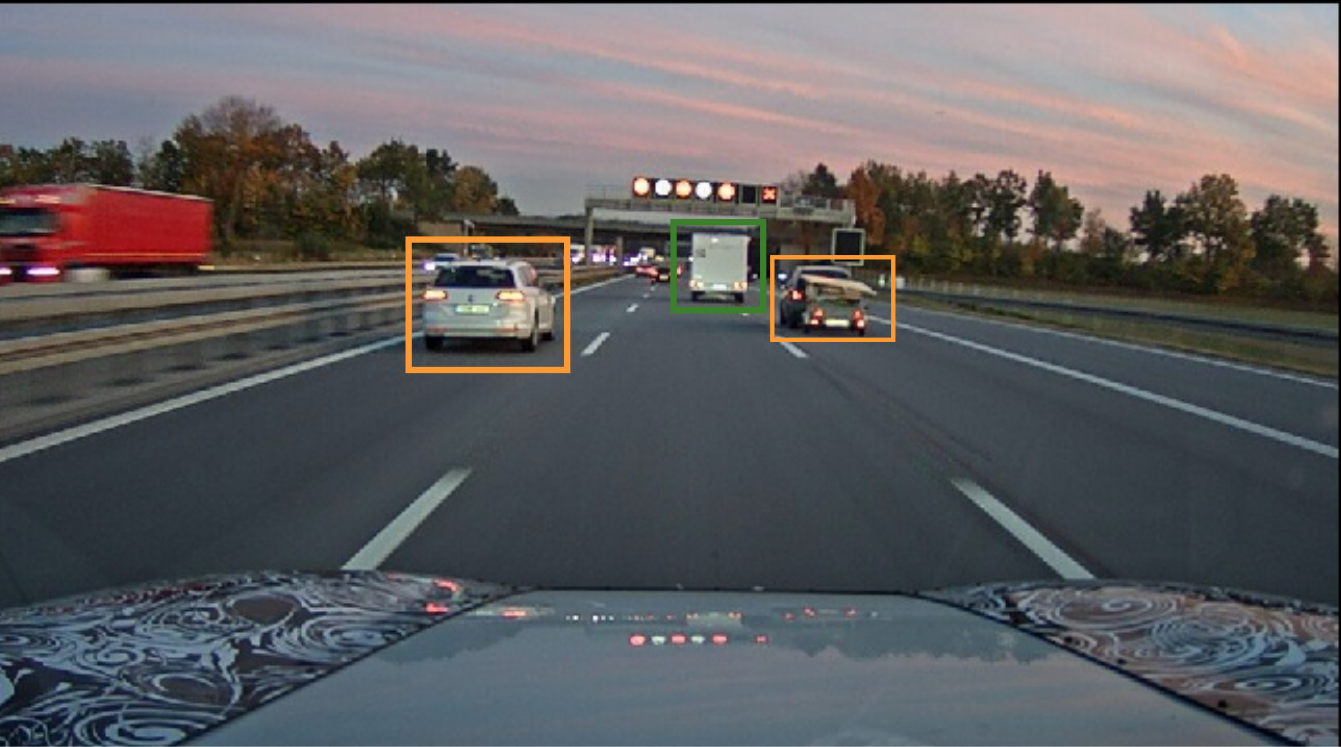}
\caption{Sample image from fleet data. Green bounding box highlights the vehicle which is doing a lane change.}
\label{fig:fleet}
\end{figure}

We use the publicly available NGSIM \cite{ngsim} dataset as well as proprietary fleet data (\cref{fig:fleet}) to demonstrate encouraging results w.r.t. state-of-the-art methods.

%\clearpage
\section{RELATED WORK}
Lane change prediction, being a fundamental building block for any autonomous driving task, is a hot topic in research and has been investigated for several years \cite{chinese_ds1, chinese_ds2, BALAL201647, ZHENG201473, 6261613}.
Picking the most informative features according to a criterion and then using ``classical" methods, like SVMs or Random Forests \cite{potsvm, DaimlerRF, inria_svm, 7576883} contributed to the core of research in lane change prediction.

Schlechtriemen et al.~\cite{DaimlerNB} analyzed the expressive power of a multitude of features and came to the conclusion that lateral distance to the lane's centerline, lateral velocity, and  relative velocity to the preceding car are the most discriminative features. They introduced two models, a Naive Bayesian approach, and a Hidden Markov  Model on top of the Naive Bayesian model, with the vanilla Naive Bayesian approach performing better.
In another work Schlechtriemen et al.~\cite{DaimlerRF} tackled the problem of predicting trajectories, where they consider lane change prediction as a helping subtask. To achieve better generalization, they fed all the available features to a random forest.

Woo et al.~\cite{potsvm} proposed a hand-crafted energy field to model the surroundings of a car for prediction with a custom SVM model.
Weidl et al.~\cite{weidl2016situation} introduced Dynamic Bayesian Networks for maneuver prediction with input features from different sensors and safety distances to the surrounding vehicles. 

A main drawback of the above approaches is the improper handling of the temporal aspect of features. A simple concatenation of features across time loses expressibility in the temporal domain, mainly due to a high degree of correlation in the features. Patel et al.~\cite{toyotasrnn}  introduced a Structural Recurrent Neural Network for this problem. Three Long Short-Term Memory (LSTM) cells handle the driving and neighbouring lanes, with inputs being the features of the surrounding vehicles in the corresponding lanes as well as features of the target.

Zeisler et al.~\cite{8317954} followed a different scheme by using raw video data instead of high-level features. Lane changes are predicted using optical flow of observed vehicles. General intention prediction is a close relative of maneuver prediction. Jain et al.~\cite{B4C} demonstrated impressive results on predicting driver intentions. The key contribution was to fuse two LSTM cells handling complementary feature spaces.

Attention mechanisms were first introduced in vision and translation tasks with outstanding performance \cite{visualattention, translationattention, Luong2015EffectiveAT}. The key idea is to guide the model towards certain points of the input, such as important image regions for visual tasks, and particularly relevant words in translation. We integrate a temporal attention mechanism into our model which cherry-picks relevant features across a sequence.

\section{PROBLEM DEFINITION}
\label{sec:problem}
Our goal is to predict lane change maneuvers of cars surrounding the ego car. %, i.e. indicate the time span from the instant a vehicle starts to cross a lane boundary till it crosses over completely to the new adjacent lane. 
Let $F_t$ be a snapshot of the scene at timestep $t$ containing $N$ vehicles. A prediction algorithm assigns a maneuver label $\{\text{left}:L, ~\text{follow}:F, ~\text{right}:R\}$ to each of the $N$ vehicles present in $F_t$. Predicting $L$ or $R$ expresses the algorithm's belief that a vehicle has started a lane change maneuver to the respective side. Predicting $F$, conversely, implies that a vehicle keeps its current lane.
\begin{figure}[!t]
\centering
\includegraphics[scale=0.6]{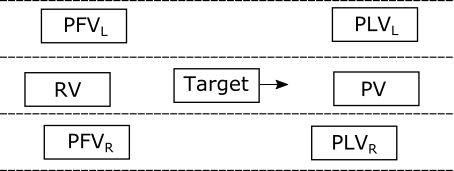}
\caption{Visualization of the dynamic environment features, direction of travel is towards the right.}
\label{fig:environment}
\end{figure}
To obtain a prediction, we use the following features for each of the $N$ cars (considered as target vehicle) in $F_t$:
\begin{itemize}
    \item \textbf{Target vehicle features}: $G^Z = (m, v_{lat}, v_{long}, a_{lat}, h)$. $m$: target's lateral distance to its lane's center line, $v_{lat}$: lateral velocity, $v_{long}$: longitudinal velocity, $a_{lat}$: lateral acceleration, and $h$: heading angle. These features are computed in Frenet coordinates. \footnote{Coordinate axis is along the target object's lane center line.}
    \item \textbf{Dynamic environment features}, i.e., features of cars surrounding the target: $G^E = (dt_X$ for $X \in \{\text{PV}, ~\text{RV}, ~\text{PLV}_L, ~\text{PLV}_R, ~\text{PFV}_L, ~\text{PFV}_R\})$, in accordance with the definition of Nie et al. \cite{7795631} (see \cref{fig:environment}). Here $dt_X$ denotes the temporal distance between the target and car $X$, i.e., the distance divided by the velocity of the trailing car.
    \item \textbf{Static environment features}: static features describe the environment type, e.g. map-based features. In the NGSIM dataset an on-/off-ramp is present, which is integrated as $G^M = (d_{\text{on}}$, $d_{\text{off}}$, $\text{lane})$. $d_{\text{on}}$, $d_{\text{off}}$ denote the distance to the nearest on-/ off-ramp respectively. $\text{lane}$ is the one hot encoding of the lane identifier.
\end{itemize}

%\clearpage
\vspace*{-0.3em}
\section{MODEL}
\vspace*{-0.2em}
We propose two kinds of recurrent networks for maneuver prediction, (a) consisting of multiple LSTM cells, and (b) an attention layer on top of that network.
We train the models in a sequence-to-sequence fashion, i.e., at every timestep $t$ an output $y_t \in \{L, F, R\}$ is generated. The input features ($G^Z$, $G^E$ and $G^M$) used for our proposed approaches are described in Section \ref{sec:problem}. 
\label{sec:LSTM}
\subsection{Long Short-Term Memory Network}
Our basic LSTM~\cite{gers1999learning} network is inspired by the work of Jain et al.~\cite{B4C}. We use three different LSTM cells ($LSTM_Z$, $LSTM_E$, $LSTM_M$), to process the feature groups ($G^Z$, $G^E$, $G^M$) respectively. This decoupling into separate LSTMs ensures that the intra-group correlation is high but the inter-group correlation is low. We use the following shorthand notation for an LSTM cell:
\begin{equation*}
    (\mathbf{h}_t^X, \mathbf{\tilde{c}}_t^X) = \texttt{LSTM}(\mathbf{X}_t, \mathbf{h}_{t-1}^X, \mathbf{\tilde{c}}_{t-1}^X)\\
\end{equation*}
where $\mathbf{X} \in \{G^Z, G^E, G^M\}$ is the input, $\mathbf{h}$ denotes the hidden state and $\mathbf{\tilde{c}}$ the memory unit.
\begin{figure}[!h]
    \centering
    \includegraphics[scale=0.5]{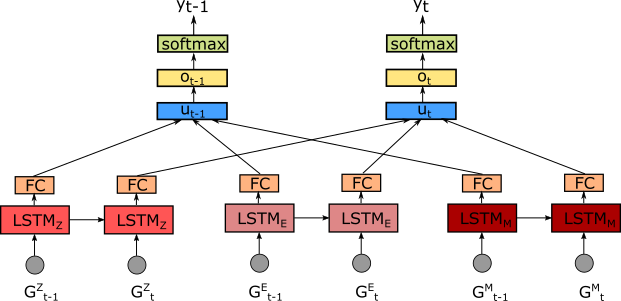}
    \caption{Visualization of the used LSTM network. Each feature category is processed by a different LSTM cell, the results are then fused. The prediction output $y$ is obtained by applying a softmax function. FC denotes a fully connected layer.}
    \label{fig:lstm}
\end{figure}
The full network can be seen in \cref{fig:lstm}. Mathematically, the fusion of these $3$ LSTMs can be formulated as:
\begin{equation}
\label{eq:lstm}
\begin{split}
\mathbf{u}_t & = \mathbf{W}_F[\texttt{concat}(\mathbf{W}_X \mathbf{h}_t^X + \mathbf{b}_X)] + \mathbf{b}_F\\
\mathbf{o}_t & = \texttt{tanh}(\mathbf{W}_{u}\cdot\mathbf{u}_t + \mathbf{b}_{u}) \\
\mathbf{y}^t & = \texttt{softmax}(\mathbf{W}_o\cdot\mathbf{o}_t + \mathbf{b}_o)
\end{split}
\end{equation}
where $\mathbf{W}$'s are the weight matrices, $\mathbf{b}$'s are bias vectors, $\mathbf{u}$ is the fusion layer, and $\mathbf{y}$ is the output layer. 
\subsection{Attention Network}
\label{sec:attention}
\begin{figure}[!t]
    \centering
    \includegraphics[scale=0.25]{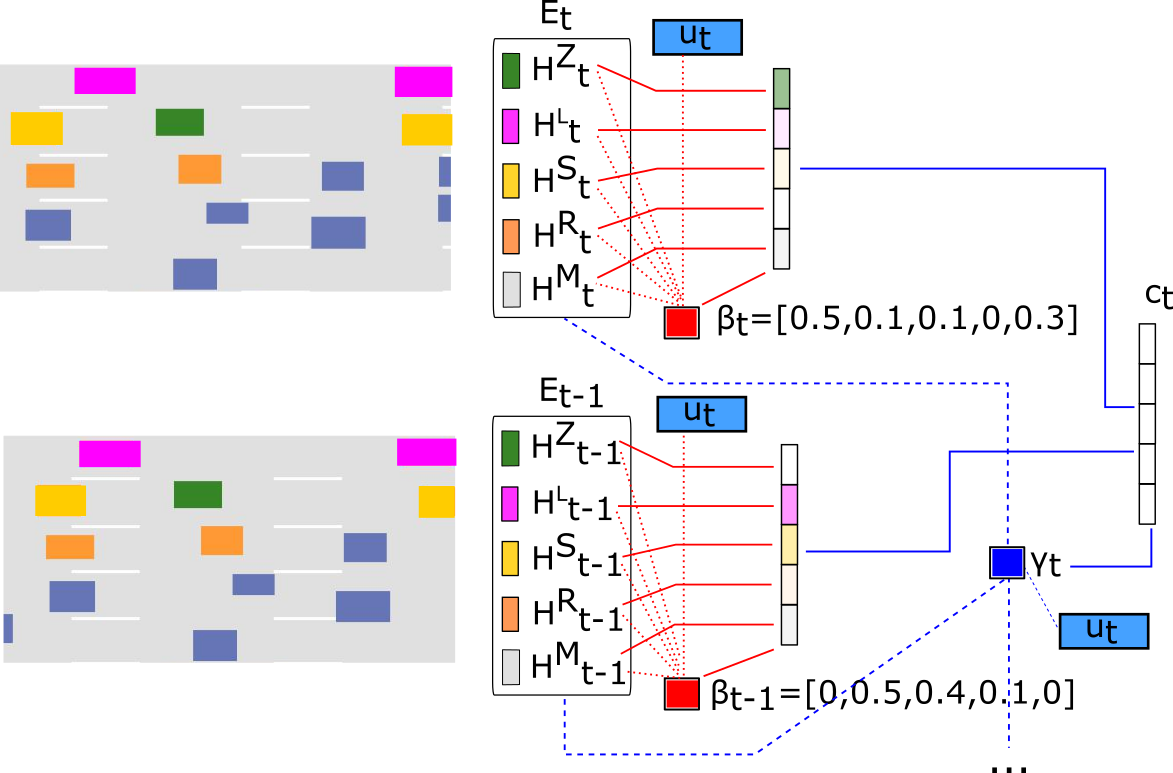}
    \caption{Attention computation for a frame at time $t$: for this, the time steps $t\!-\!l, \ldots, t$ are considered. Scenes for times $t$ and $t-1$ are drawn in vertical order. The embeddings $\mathbf{E}_t$, $\mathbf{E}_{t-1}$ are shown next to it. Using the layer $\mathbf{u}_t$ as query, the attention weights $\mathbf{\beta}_t$, $\mathbf{\beta}_{t-1}$ are calculated, with which the embeddings are then scaled category-wise. Again using $\mathbf{u}_t$ as query, the attention weights $\mathbf{\gamma}_t$ are calculated, showing the importance of the different time steps. The weighted sum of the scaled embeddings  w.r.t. to $\mathbf{\gamma}_t$ makes up the context vector $\mathbf{c}_t$.}
    \label{fig:attention}
\end{figure}

The idea behind an attention mechanism is to model \emph{selective focus}, i.e.,  on certain parts of the input.
It mainly consists of a function processing a \textit{key}~(K) and \textit{query}~(Q) to obtain a context vector, which is the accumulation of multiple keys, weighted by their importance w.r.t. the query. We employ two kinds of attention mechanisms, (a) attention over previous time steps, i.e., self-attention \cite{selfattention}, and (b) attention over different feature groups. 
%In a first step, feature importances are determined by using the current hidden state of the network, and then the same is done for previous time steps. 
As opposed to traditional attention approaches, the features we use lie in different spaces and have different modalities. We do not accumulate them, but only change their magnitude in accordance with the weighting, and then accumulate these feature vectors over the time steps; see \cref{fig:attention} for an intuitive visualization.
%To the best of our knowledge this is the first use of an attention mechanism which fuses diverse input modalities, see \cref{fig:attention} for an intuitive visualization.

We again partition the features into categories, but with a finer granularity than in \cref{sec:problem}, \textit{viz}. $H^Z = G^Z$, $H^S = [dt_{\text{PV}}, dt_{\text{RV}}]$, $H^L = [dt_{\text{PLV}_L}, dt_{\text{PFV}_L}]$, $H^R = [dt_{\text{PLV}_R}, dt_{\text{PFV}_R}]$ and $H^M=G^M$.
The attention function $\Psi \colon \mathbb{R}^d \to \mathbb{R}$ is given by:
\begin{equation}
\begin{split}
\Psi(\mathbf{W}, \mathbf{v}, \texttt{Q}, \texttt{K}) & = \mathbf{v}^T \texttt{tanh}(\mathbf{W}[\texttt{Q; K}])
\end{split}
\end{equation}
For time step $t$ in all calls of $\Psi$, layer $\mathbf{u}_t$ serves as query.
Let $T = \{t\!-\!l, \ldots, t\}$ be the time steps used for self-attention; we have used $l=20$ in our experiments. For each $i \in T$ the feature categories are embedded into a higher-dimensional space, and the importances of each feature category, $\mathbf{\beta}^X_i$, as well as each time step as a whole, $\mathbf{\gamma}_i$, are determined. Let $C = \{H^Z, H^L, H^S, H^R, H^M\}$:
\begin{equation}
\begin{split}
\mathbf{E}_i^X & = \mathbf{W}_{EX} \mathbf{X}_i + \mathbf{b}_{EX}\\
\mathbf{\beta}^X_i & = \Psi(\mathbf{W}_{FX}, \mathbf{v}_{FX}, \mathbf{u}_t, \mathbf{E}_i^X)\\
\mathbf{\gamma}_i & = \Psi(\mathbf{W}_{Time}, \mathbf{v}_{Time}, \mathbf{u}_t, \texttt{concat}([\mathbf{E}_i^{X}])\\
\mathbf{\beta}_i &= \texttt{softmax}([\texttt{concat}(\mathbf{\beta}^{X}_i)])
\end{split}
\end{equation}
where $X \in C$, $i \in T$.
Eventually, the feature categories are scaled by $\mathbf{\beta}^X_i$ and the weighted sum is calculated over all time steps. The resulting context vector is appended to the fusion layer and the computation follows \cref{eq:lstm}.
\begin{equation}
\begin{split}
\mathbf{\gamma}_t &= \texttt{softmax}([\texttt{concat}( \mathbf{\gamma}_i)]) \\
\mathbf{c}_t & = \sum_{i \in T}\mathbf{\gamma}_i \texttt{concat}([\mathbf{\beta}_i \mathbf{X}_i ])\\
\mathbf{u}_t &= [\mathbf{u}_t; \mathbf{c}_t] \\
\end{split}
\end{equation}
\subsubsection{Visualization of Attention}
\label{sec:vis}
Apart from improved performance, another large benefit of attention is its interpretability. Traditionally, simply the magnitude of the attention weights, which are used in the calculation of the weighted mean, is shown \cite{translationattention}. Here though, due to the different scales and dimensions of the feature categories, this does not necessarily lead to expected results. Instead, we calculate the derivative of the predicted class by the attention weights $\mathbf{\beta}_i^X$ and $\mathbf{\gamma}_i$, summing over all time steps. This derivative denotes the contribution of category $X$ to the resulting prediction, even providing the information whether this contribution is positive or negative.
\subsection{Training Scheme}
\label{sec:training}
As proposed in \cite{B4C}, we employ an exponentially growing loss to encourage early predictions. The used Softmax loss is weighted with  $\alpha\!\cdot\!w_t\!\cdot\!\exp(-T)$, where at time $t$ a lane change is imminent in the next $T$ seconds.\footnote{Exponential weighting of the loss function is not done for the fleet data, as the human labels are error-free.}
We choose $\alpha$ s.t. the average value of $\alpha\!\cdot\!\exp(-T)$ over all frames of each lane change maneuver equals $1$. For a given maneuver at time $t$, $w_t$ is inversely proportional to that maneuver's global size in training data.

As noted by Schlechtriemen et al. \cite{DaimlerNB}, simple scenarios cover a majority of lane changes, and a relatively good prediction can already be achieved by using a small subset of features from $G^Z$. To tackle this imbalance and induce a meaningful gradient flow for the attention in all cases, we introduce a dropout layer in between layer $\mathbf{u}$ and $\mathbf{o}$, i.e. 
\begin{equation*}
 \texttt{Dropout} = [\mathbf{W}_{Drop, Fusion};\texttt{ }\mathbf{W}_{Drop, c}]\cdot[\mathbf{u}_t;\texttt{ }\mathbf{c}_t] + \mathbf{b}_{drop}   
\end{equation*}
With a probability $p = 0.33$, $\mathbf{W}_{Drop, Fusion}$ and $\mathbf{W}_{Drop, c}$ are set to 0 independently, forcing the model to rely solely on its recurrent architecture or attention.

%\clearpage
\section{Datasets and Evaluation}
\subsection{Datasets}
\noindent\textbf{NGSIM}: The Next Generation Simulation (NGSIM)~\cite{ngsim} project consists of four publicly available traffic data sets. We use the US Highway 101 dataset (US-101) and Interstate 80 Freeway dataset (I-80). Data is captured from a bird's-eye view of the highway with a static camera, and vehicle-related high-level features are extracted from it. The datasets contain measurements at $10Hz$. After removing noisy trajectories, $3184$ lane changes are observed. 
%Since, the lane coordinates are missing in the dataset, we have estimated the lane coordinates using linear least squares regression on the observed coordinates of lane changes after resampling the dataset at $25Hz$.

% \begin{figure}[!h]
% \centering
% \includegraphics[scale=0.35]{img/us101}
% \caption{Sketch of the US-101 dataset (proportions are not to the actual scale). Note that the I-80 dataset has no off-ramp.}
% \label{fig:us101}
% \end{figure}

\noindent\textbf{Fleet data:} The fleet data comes from the fused perception model of in-production cars. This data is captured at $25Hz$ w.r.t. to a moving ego car equipped with several camera and radar sensors to give a complete $360^\circ$ view. 830 lane changes are recorded.

\subsection{Metrics}
\label{subsec:metrics}
A wide variety of metrics is used to measure the performance of lane change prediction algorithms. Predominantly they are inspired by information retrieval %i.e. Precision and Recall. Besides, these metrics
and are computed treating each timestep independently of the other.
%\vspace*{-0.5em}
\begin{itemize}
    \item\textbf{Accuracy}: percentage of timesteps correctly classified.
\end{itemize}
Jain et al. \cite{B4C} introduced a first version of the following maneuver-based metrics: 
\begin{itemize}
    \item\textbf{Precision}: percentage of true predictions w.r.t. total number of maneuver predictions.
    \item\textbf{Recall}: percentage of true predictions w.r.t. total number of maneuvers.
    \item\textbf{Time to Maneuver (TTM)}: the interval between the time of prediction and the actual start of the maneuver in the ground truth. 
\end{itemize}
For evaluation, we combine Precision and Recall into the F1 score and refer to the metrics introduced by Jain et. al \cite{B4C} by their associated group's name Brain4Cars (B4C).
\begin{figure}[!h]
    \centering
    \includegraphics[scale=0.8]{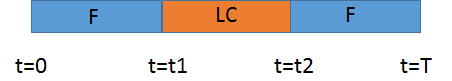}
    \caption{Maneuver labels are continuous events. Here `F' denotes follow from [0,$t_1$] and [$t_2$, T], and `LC' denotes a lane change event [$t_1$, $t_2$]. }
    \label{fig:lc_event}
\end{figure}

%Since, the labels are event-wise continuous (\cref{fig:lc_event}), we define the following \textit{event-wise} metrics:
%\begin{itemize}
%    \item\textbf{Precision:} whatever is predicted is it the correct maneuver type. Higher number of false positives implies lower precision.
%$P = \frac{1}{M}\sum_{i=1}^{M}\frac{| s_i \cap g |}{|s_i|}$
%    \item\textbf{Recall:} are all ground truth maneuvers correctly predicted.
%$R = \frac{1}{N}\sum_{i=1}^{N}\frac{| g_i \cap s |}{|g_i|}$
%\end{itemize}
%Note that $M$ need not be equal to $N$ (\cref{fig:maneuver_pr}).
%\begin{figure}[ht]
%    \centering
%    \includegraphics[scale=0.4]{img/maneuver_PR.png}
%    \caption{Our precision recall inspired from \cite{NB13}. The accompanying black arrow’s direction indicates how the chunks are created for each measure. The number of prediction events $M$ is 5, whereas number of ground truth events $N$ is 3.}
%    \label{fig:maneuver_pr}
%\end{figure}

\begin{figure}[!t]
    \centering
    \includegraphics[scale=0.8]{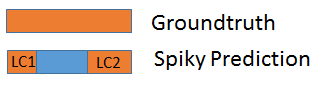}
    \caption{If a given ground truth event has multiple corresponding prediction events then for overlap $LC1$ is used. The comfort related metrics for this event will be delay=0, overlap=20\%, frequency=2, miss=0}
    \label{fig:maneuver_pr}
\end{figure}

The ground truth labels are event-wise continuous (\cref{fig:lc_event}). The information retrieval metrics, however, do not reflect this event-wise nature or what the driver experiences in the car. The car's controller usually reacts to the first prediction event (\cref{fig:maneuver_pr}). If the prediction is discontinuous then this causes discomfort to the driver (stop-and-go function). In addition, the prediction event should be as early as possible w.r.t. the ground truth, and the earlier the prediction the higher is the comfort. In order to reflect such comfort-related behaviour we propose the following \textit{event-wise} metrics:
%\vspace*{-0.5em}
\begin{itemize}
    \item\textbf{Delay:} delay (measured in seconds) in prediction w.r.t. the ground truth label. If prediction is perfectly aligned with the ground truth then delay is $0$.
    \item\textbf{Overlap:} for a given ground truth event the percentage of overlap for the earliest maneuver predicted. The higher the overlap, the smoother is the controller's reaction.
    \item\textbf{Frequency:} number of times a maneuver event is predicted per ground truth event. For the `follow' event this indicates the false positive rate (FPR).
    \item\textbf{Miss:} number of lane changes completely missed. The higher the number of misses, the higher is the discomfort, as the driver has to intervene.
\end{itemize}

\subsection{Labeling}
\begin{figure}[!t]
    \centering
    \includegraphics[scale=0.45]{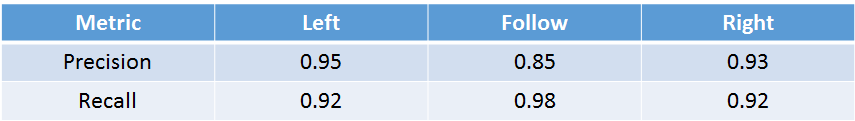}
    \caption{Comparing human vs human labels: Precision and Recall of ground truth lane changes on proprietary fleet data. Each trace was labeled by 3 different humans. Metrics are computed in a ``$1$ vs rest fashion" then averaged. This shows that humans agree on almost all lane changes, but there is a slight disagreement on when the maneuver actually starts, i.e notion of intention of maneuver varies across humans. }
    \label{fig:gt_stats_bmw}
\end{figure}

The perception of the precise moment when a lane change starts differs from person to person, see \cref{fig:gt_stats_bmw}. Therefore, manually labeling lane changes in fleet data gives us a hint at the intention. However, automatic labeling is useful in the case of NGSIM due to a similar time span of lane changes. Thus, like \cite{DaimlerNB} we have used a $3$-second criterion, before the target's lane assignment changes, to label a lane change.
%\footnote{NGSIM's train and test set will be released for a transparent comparison}%
Though human labeling is precise and error-free, it is time-consuming and expensive. Intelligent automatic labeling can be slightly imprecise, but on the other hand, is quicker and might prove to be better for deep models, which could pick up on fine cues imperceptible to humans to achieve a better performance.
%Labeling lane changes is a crucial and tedious task, as the perceived precise moment when a lane change starts differs from person to person, see \cref{fig:gt_stats_bmw}. An alternative to human labeling is automatically labeling a window of $T$ seconds before a lane change with the corresponding maneuver type. Both methods have their pros and cons: Human labeling is more precise and does not contain any wrong labels, but it is costly and creates a cap on the achievable performance. On the other hand, automatic labeling is cheap and might prove more useful for powerful deep models, which could pick up on fine cues imperceptible to humans to achieve a better performance. However, using a fixed window size results in some wrong labels where the maneuver may not have not started. As our fleet data is labeled by humans and we use automatic labels on NGSIM with $T = 3s$. Another interesting contribution of this work is the comparison of the same methods under different labeling schemes.

%Measurements in the above datasets are often noisy, but previous works have not clearly defined which trajectories they use and which they discard. Therefore, to have a transparent comparison we will open source our training and test set as well as the automatic labels for NGSIM only.%\footnote{code for models used for comparison and metrics will also be released}

%\clearpage
\section{RESULTS}
We denote our two proposed recurrent methods in \cref{sec:LSTM} as \textit{LSTM-E} (extended LSTM) and \textit{LSTM-A} (extended LSTM with attention). For both a hidden size of 128 is used. We implement state-of-the-art baselines to demonstrate better performance of our proposed methods.

\subsection{Baseline-Methods}
\label{subsec:baselines}
\noindent\textbf{Frame-based:} Features from a single timestep are used.
\begin{itemize}
\item{Random Forest (RF) \cite{DaimlerRF}}: The concatenated features ($G^Z, G^E, G^M$) serve as input.
\item Naive Bayes (NB) \cite{DaimlerNB}: The features $m, v_{lat}$ and relative velocity to preceding car are used.
\end{itemize}
\noindent\textbf{Sequence-based:}
\begin{itemize}
\setlength\itemsep{-0.7em}
\item{Structural RNN (SRNN) \cite{toyotasrnn}}: The SRNN consists of three different LSTM cells which cover the target, left, and right lane respectively. To each LSTM cell the features $Q$ of three vehicles are given, \textit{viz.} those of the two neighbors of the target car ($\text{PV}$ - $\text{RV}$ / $\text{PLV}_L$ - $\text{PFV}_L$ / $\text{PLV}_R$ - $\text{PFV}_R$) and the target car itself. $Q$ consists of absolute world coordinates, lateral and longitudinal velocity, heading angle, and number of lanes to the left and right. The output of the three LSTM cells is passed on to another LSTM cell, which eventually outputs the prediction.\\
\item{Vanilla LSTM (LSTM)}: Vanilla LSTM consisting of a single cell with the concatenated features $(G^Z, G^E, G^M)$.\\
\end{itemize}

\subsection{Quantitative Results}
\label{subsec:quantres}
\cref{tab:multicol} shows the results of all tested methods w.r.t. all metrics on the NGSIM and fleet dataset. As can be seen, due to the diversity of the evaluation metrics, some methods excel or fail in different categories. Sequence-based methods easily outperform frame-based methods since the latter carry no information regarding the sequence history. Among sequence-based methods, our three recurrent models, LSTM, LSTM-E, and LSTM-A come out on top (refer to the `Total Rank' column in the table). 

On the NGSIM dataset, the LSTM network with attention is the best-performing method. It has the lowest delay while predicting lane changes, a lower false positive rate during `follow', and a good continuous prediction indicated by `Overlap'. On our fleet data LSTM-A finishes second. This is mainly due to the sparsity of the dynamic environment features $G^E$ in fleet data. Thus, the prediction falls back to the target features $G^Z$, as these are the most discriminative features, and the performance is similar to vanilla LSTM.
%A plausible reason for this is the simplistic structure of the dataset, as the scenes tend to be less crowded, and thus do not utilize the full power of a complex attention mechanism.

%While both mentioned labeling schemes (human and automatic) clearly have their advantages and disadvantages, it is possible that methods predict "too well", contrasting the human ground truth and decreasing the numerical performance

\subsection{Qualitative Results}
\label{subsec:qualres}
As can be seen from \cref{tab:multicol}, the performance of some methods is relatively similar. Analyzing and interpreting a few critical corner cases will help in assessing the performance and give us clarity about the advantage of an attention mechanism. These critical corner cases are not present in the data. They were created by translating around the existing trajectories w.r.t. their position in the scene, and thus remain realistic.
For the fleet data we do not have the static data recording, but instead a moving ego car (drawn in red) from which the measurements of the scene are obtained.

Two types of visualizations are used: (a) a snapshot visualization of a single frame, and (b) a visualization of the temporal development of a scene. The first consists of a single image, showing the ground truth and prediction of a single algorithm for that frame, as well as the attention visualization for the five feature categories. For better readability the categories $H^Z$, $H^S$, $H^L$, $H^R$ and $H^M$ are denoted by \emph{Target}, \emph{Same}, \emph{Left}, \emph{Right}, and \emph{Street}. (b) is the concatenation of several frames spanning a certain amount of time, along with the prediction of different algorithms.
%We start with some sample scenes pointing out the behavior and contribution of attention in these, the visualization and with that connected use of the term "contribution" of the attention categories is explained in \cref{sec:vis}.
\begin{figure}[!t]
\includegraphics[scale=0.3]{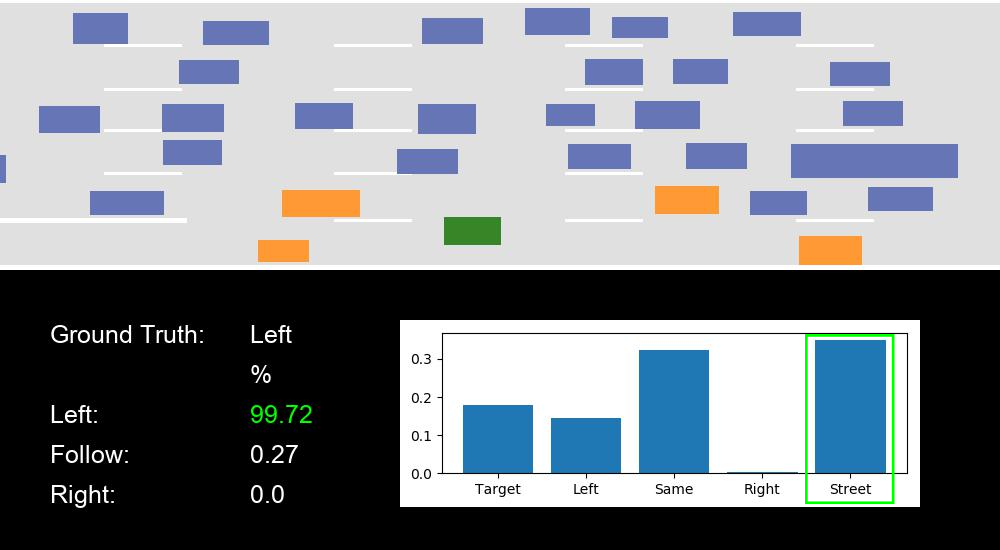}
\caption{The target car (green) is executing a lane change to the left from the auxiliary lane. As changing lanes after an on-ramp is the usual and expected behavior, \emph{Street} has a high positive contribution for predicting $L$.}
\label{fig:onramp}
\end{figure}

\begin{figure*}[!t]
\subfigure[The target (green) rapidly approaches $\text{PV}$, and humans can anticipate a forthcoming lane change. The same reasoning is done by the attention mechanism, \emph{Same} has a strong negative contribution to the prediction $F$, and positive contribution to the eventual predicton of $L$.]{
\includegraphics[width=.19\paperwidth]{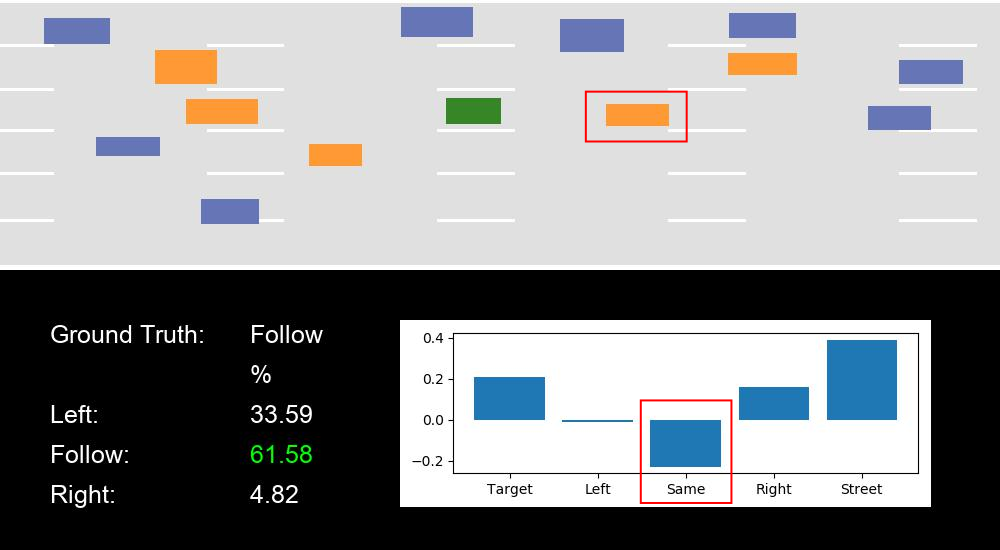}
\includegraphics[width=.19\paperwidth]{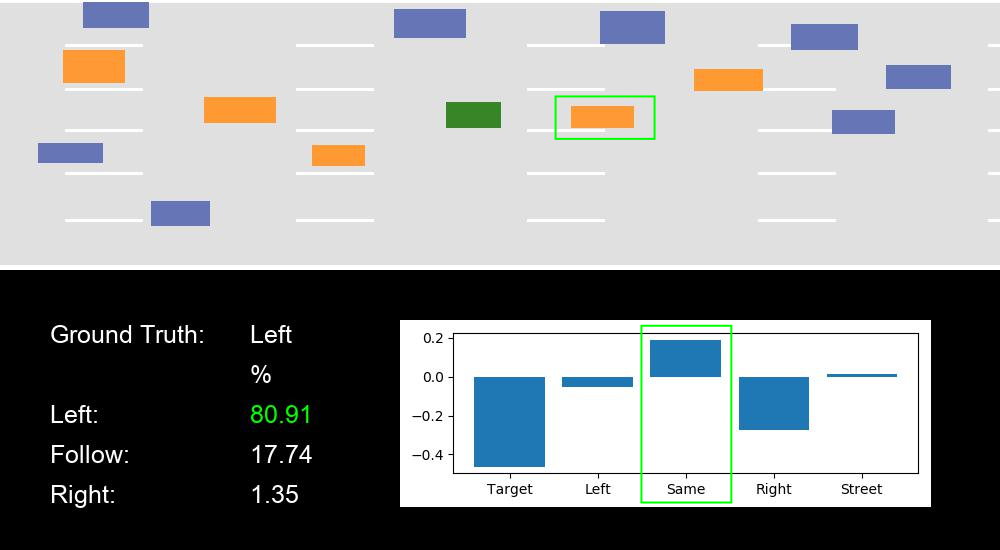}
}
\subfigure[The contribution of \emph{Same} is correctly reversed w.r.t. a.]
{
\includegraphics[width=.19\paperwidth]{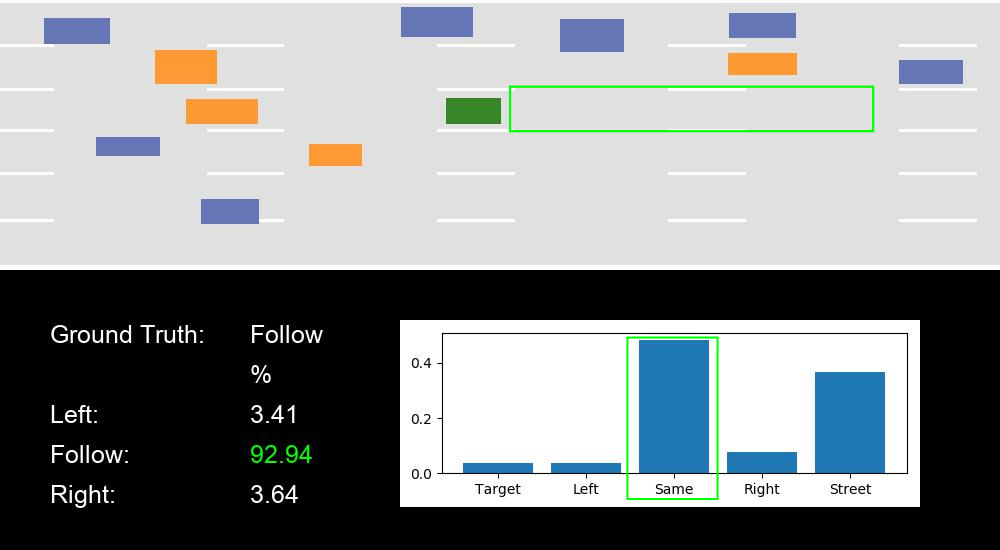}
\includegraphics[width=.19\paperwidth]{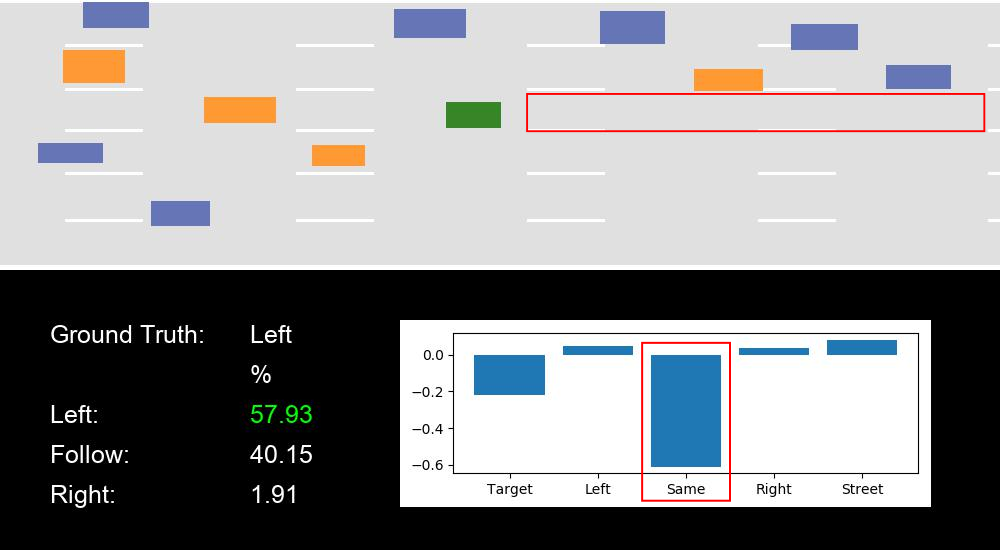}
}
\caption{
Two scenes of a lane change to the left are shown. Scene (a) and (b) stem from the same real scene and differ only in the placement of $\text{PV}$, which is close to the target in (a) and missing in (b).
In the first image, respectively, $F$ is predicted, in the second $L$.  }
\label{fig:PV}
\end{figure*}

\cref{fig:onramp} and \cref{fig:PV} show the influence of attention on the network's decision making, highlighting its correct and intuitive contribution.

\begin{figure*}[!t]
\centering
\subfigure[Visualization of a lane change to the right as recorded from a fleet car (\cref{fig:fleet} shows the front-camera image). LSTM-A predicts first, followed by Random Forest, and LSTM-E.]{
  \includegraphics[width=\textwidth]{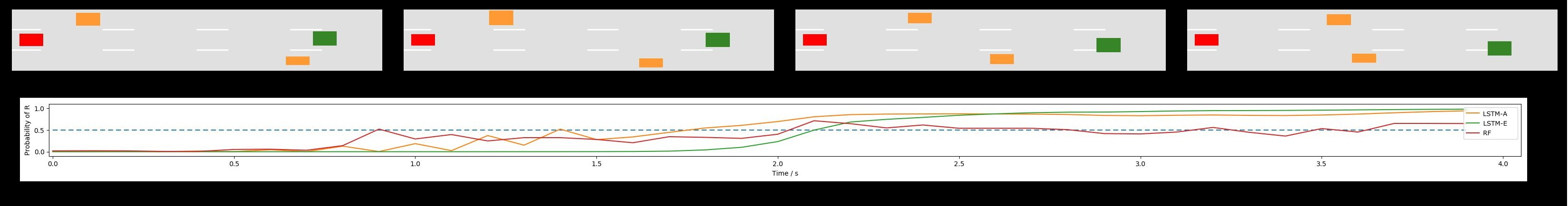}
}
\subfigure[Depiction of a ``fake" lane change to the left. 
The target car (green) starts moving towards its left lane boundary. The lane change is impossible because it is blocked by neighbouring cars (orange). Despite this, Random Forest is ``fooled" quickly and predicts a `left' lane change, whereas the recurrent networks correctly predict `follow' and have a false prediction only towards the end. Note that also an attention mechanism cannot fully prevent a false lane change behavior, as the relation to surrounding cars is learned and not hard-coded. Once a strong lateral movement is observed, one has to consider the possibility of a coming lane change, independent of the current road situation. However, in most of such cases the attention mechanism can prevent a false lane change prediction.]{
  \includegraphics[width=\textwidth]{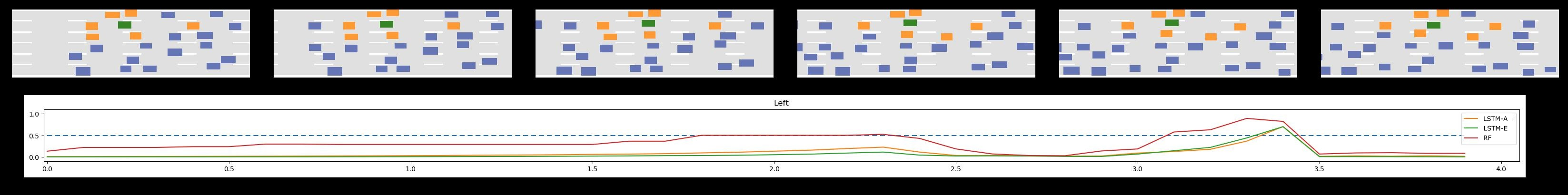}
}
\caption{Temporal visualization of two lane change maneuvers.}
\label{fake_left}
\end{figure*}

\cref{fake_left} shows the temporal development of two scenes while plotting the output of three algorithms - RF, LSTM-E, and LSTM-A.
Overall a superior performance of the recurrent models, especially LSTM-A, can be observed.
% todo - add this to text - Dataset included about 1000 lane changes. Split dataset into train, validation and test, results on test are shown, so far only 1 run. This is just a first version, e.g. B4C metric has to be adjusted for this.

% \vspace*{-10cm}

%\vspace*{-3em}
\begin{table*}[!b]
%\tiny{
\caption{Comparison and ranking of different methods on different metrics for NGSIM (top) and fleet data (bottom). Total Rank is the ordering on the average of an algorithm's rank per column, the lower, the better. The rank of an algorithm for a particular column is the index it occupies after sorting that column.}
%}
\begin{center}
%\subcaption*{NGSIM dataset}
\begin{tabular}{c|c|c|c|c|c|c|c|c|c|c|c|c|c}
    & \multicolumn{3}{c|}{Frame-based} & \multicolumn{9}{c|}{Maneuver-based} & \\
   \cline{2-13}
     & \multicolumn{3}{c|}{Accuracy} & \multicolumn{2}{c|}{B4C} & \multicolumn{7}{c|}{Proposed} & \\
   \cline{2-13}
    & L & F & R & F1 & TTM & \multicolumn{3}{c|}{L} & \multicolumn{1}{c|}{F} & \multicolumn{3}{c|}{R} & \\
   \cline{2-13}
    Algorithm & &  &   &  &   &  Miss & Delay &  Over & Freq & Miss & Delay & Over  & Total Rank\\
    \hhline{=|=|=|=|=|=|=|=|=|=|=|=|=|=}
NB & 0.715 & 0.886 & \textbf{0.679} & 0.0 & 0.0 & \textbf{0.002} & 0.269 & 0.64 & 7.271 & \textbf{0.003} & 0.295 & \textbf{0.595} & 5\\

RF & 0.744 & 0.938 & 0.67 & 0.767 & 0.965 & 0.003 & 0.231 & 0.636 & 7.231  & 0.006 & \textbf{0.269} & 0.513 & 4\\

SRNN & 0.555 & 0.905 & 0.475 & 0.475 & 1.337 & 0.212 & 0.22 & 0.521 & 6.686 & 0.121 & 0.355 & 0.4  & 6\\

LSTM & 0.772 & 0.958 & 0.634 & \textbf{0.822} & 1.086 & \textbf{0.002} & 0.215 & 0.671 & 4.749 & 0.007 & 0.341 & 0.525 & 2\\

LSTM-E & 0.759 & \textbf{0.962} & 0.603 & 0.813 & 1.093 & 0.003 & 0.228 & 0.666 & \textbf{4.327} & 0.012 & 0.363 & 0.493 & 3\\

LSTM-A & \textbf{0.784} & 0.951 & 0.662 & 0.802 & \textbf{1.138} & 0.003 & \textbf{0.207} & \textbf{0.694} & 5.34 & 0.01 & 0.306 & 0.547  & \textbf{1}
\end{tabular}

\bigskip
%\subcaption*{Fleet data}
\begin{tabular}{c|c|c|c|c|c|c|c|c|c|c|c|c|c}
    & \multicolumn{3}{c|}{Frame-based} & \multicolumn{9}{c|}{Maneuver-based} & \\
   \cline{2-13}
     & \multicolumn{3}{c|}{Accuracy} & \multicolumn{2}{c|}{B4C} & \multicolumn{7}{c|}{Proposed} & \\
   \cline{2-13}
    & L & F & R & F1 & TTM & \multicolumn{3}{c|}{L} & \multicolumn{1}{c|}{F} & \multicolumn{3}{c|}{R} & \\
   \cline{2-13}
    Algorithm & &  &   &  &   &  Miss & Delay &  Over & Freq & Miss & Delay & Over  & Total Rank\\
    \hhline{=|=|=|=|=|=|=|=|=|=|=|=|=|=}

NB & 0.767 & 0.773 & 0.879 & 0.0 & 0.0 & 0.062 & 0.083 & 0.622 & 3.461 & \textbf{0.025} & \textbf{0.04} & 0.759 & 5\\

RF & 0.843 & \textbf{0.886} & 0.854 & 0.321 & \textbf{1.071} & 0.064 & \textbf{0.06} & 0.643 & 3.515 & 0.038 & 0.056 & 0.709 & 4\\

SRNN & 0.868 & 0.834 & 0.852 & 0.275 & 0.691 & 0.054 & 0.087 & 0.782 & 2.28 & 0.05 & 0.09 & 0.771  & 6\\

LSTM & \textbf{0.904} & 0.883 & \textbf{0.926} & \textbf{0.597} & 0.795 & \textbf{0.038} & 0.062 & \textbf{0.822} & \textbf{1.937} & 0.027 & 0.047 & \textbf{0.859} & \textbf{1}\\

LSTM-E & 0.898 & 0.884 & 0.916 & 0.487 & 0.757 & 0.052 & 0.061 & 0.807 & 2.247 & 0.043 & 0.048 & 0.852 & 3\\

LSTM-A & 0.899 & 0.876 & 0.924 & 0.432 & 0.622 & 0.042 & 0.065 &  0.814 & 2.238 & 0.045 & 0.045 & \textbf{0.859} & 2\\
\end{tabular}
\end{center}
\label{tab:multicol}
%}
\end{table*}

\section{CONCLUSIONS}
We have proposed an LSTM network with an attention mechanism for lane change prediction, which performs better than existing methods w.r.t. to different evaluation schemes. This is the first work applying such a model to this field, which tackles both prediction quality and understandability. We have also proposed new event-wise metrics catering to driver's comfort. 
%Decoupling of LSTM cells with a smart embedding of different feature groups has avoided the need to gather innumerable corner cases for the training data. 
Results on a public dataset as well as fleet data clearly indicate a high level of comfort, in terms of earliness in prediction, false positive, and miss rate, with our proposed methods for the driver.
Moreover, with visual analysis of critical cases we have demonstrated the effectiveness of using attention. In the future, analyzing fleet data with complex scenes using our attention mechanism can shine light on circumventing critical cases for fully autonomous driving. Such understandable mechanisms are helpful in diagnosing and minimizing accidents. This can eventually lead to improved path planning algorithms.
\clearpage

%\addtolength{\textheight}{-20cm}   % This command serves to balance the column lengths

%%%%%%%%%%%%%%%%%%%%%%%%%%%%%%%%%%%%%%%%%%%%%%%%%%%%%%%%%%%%%%%%%%%%%%%%%%%%%%%%

%%%%%%%%%%%%%%%%%%%%%%%%%%%%%%%%%%%%%%%%%%%%%%%%%%%%%%%%%%%%%%%%%%%%%%%%%%%%%%%%

%%%%%%%%%%%%%%%%%%%%%%%%%%%%%%%%%%%%%%%%%%%%%%%%%%%%%%%%%%%%%%%%%%%%%%%%%%%%%%%%
%\vspace*{-5cm}

% \let\OLDthebibliography\thebibliography
% \renewcommand\thebibliography[1]{
%   \OLDthebibliography{#1}
%   % \setlength{\parskip}{0pt}
%   \setlength{\itemsep}{0.2pt}
% }

\clearpage
\bibliographystyle{IEEEtran}
%\bibliography{bibliography}

\end{document}